\documentclass[letterpaper]{article} % DO NOT CHANGE THIS
\usepackage{aaai25}  % DO NOT CHANGE THIS
\usepackage{times}  % DO NOT CHANGE THIS
\usepackage{helvet}  % DO NOT CHANGE THIS
\usepackage{courier}  % DO NOT CHANGE THIS
\usepackage[hyphens]{url}  % DO NOT CHANGE THIS
\usepackage{graphicx} % DO NOT CHANGE THIS
\usepackage{natbib}  % DO NOT CHANGE THIS AND DO NOT ADD ANY OPTIONS TO IT
\usepackage{caption} % DO NOT CHANGE THIS AND DO NOT ADD ANY OPTIONS TO IT
\usepackage{algorithm}
\usepackage{algorithmic}
\usepackage{amsmath}
\usepackage{bm}
\usepackage{amssymb}
\usepackage{amsfonts}
\usepackage{cite} %
\usepackage{csquotes} %
\usepackage{physics}
\usepackage{booktabs}
\usepackage{multirow}
\usepackage{xcolor} % For text colors
\usepackage{threeparttable}
\usepackage{amsfonts}
\usepackage{arydshln}
\usepackage{threeparttable}
\usepackage{newfloat}
\usepackage{listings}
\DeclareCaptionStyle{ruled}{labelfont=normalfont,labelsep=colon,strut=off} % DO NOT CHANGE THIS
\floatstyle{ruled}
\newfloat{listing}{tb}{lst}{}
\floatname{listing}{Listing}

\title{VQ4DiT: Efficient Post-Training Vector Quantization for Diffusion Transformers}
\author{
    Juncan Deng\textsuperscript{\rm 1}\equalcontrib, Shuaiting Li\textsuperscript{\rm 1}\equalcontrib, Zeyu Wang\textsuperscript{1}, Hong Gu\textsuperscript{\rm 2}, Kedong Xu\textsuperscript{\rm 2}, Kejie Huang\textsuperscript{\rm 1}\\
}
\affiliations{
    \textsuperscript{\rm 1}Zhejiang University,  \textsuperscript{\rm 2}vivo Mobile Communication Co., Ltd \\
    \{dengjuncan, list, wangzeyu2020, huangkejie\}@zju.edu.cn, \\
    \{guhong, xukedong\}@vivo.com \\
}

\usepackage{bibentry}

\begin{document}

\maketitle

\begin{abstract}
The Diffusion Transformers Models (DiTs) have transitioned the network architecture from traditional UNets to transformers, demonstrating exceptional capabilities in image generation. Although DiTs have been widely applied to high-definition video generation tasks, their large parameter size hinders inference on edge devices. Vector quantization (VQ) can decompose model weight into a codebook and assignments, allowing extreme weight quantization and significantly reducing memory usage. In this paper, we propose VQ4DiT, a fast post-training vector quantization method for DiTs. We found that traditional VQ methods calibrate only the codebook without calibrating the assignments. This leads to weight sub-vectors being incorrectly assigned to the same assignment, providing inconsistent gradients to the codebook and resulting in a suboptimal result. To address this challenge, VQ4DiT calculates the candidate assignment set for each weight sub-vector based on Euclidean distance and reconstructs the sub-vector based on the weighted average. Then, using the zero-data and block-wise calibration method, the optimal assignment from the set is efficiently selected while calibrating the codebook. VQ4DiT quantizes a DiT XL/2 model on a single NVIDIA A100 GPU within 20 minutes to 5 hours depending on the different quantization settings. Experiments show that VQ4DiT establishes a new state-of-the-art in model size and performance trade-offs, quantizing weights to 2-bit precision while retaining acceptable image generation quality.
\end{abstract}

\section{Introduction}
Advancements in pre-trained text-to-image diffusion models \cite{ho2020denoising, ho2022cascaded, ramesh2022hierarchical, rombach2022high, saharia2022photorealistic} have facilitated the successful generation of images that are both complex and highly faithful to the input conditions. Recently, Diffusion Transformers Models (DiTs)  \cite{peebles2023scalable} have garnered significant attention due to their superior performance, with OpenAI's SoRA \cite{sora} being one of the most prominent applications. DiTs are constructed by sequentially stacking multiple transformer blocks. This architectural design leverages the scaling properties of transformers \cite{carion2020end, touvron2021training, xie2021segformer, liu2021swin}, allowing for more flexible parameter expansion to achieve enhanced performance. Compared to other UNet-based diffusion models, DiTs have demonstrated the ability to generate higher-quality images while having more parameters.

Deploying DiTs can be costly due to their large number of parameters and high computational complexity, which is similar to the challenges encountered with Large Language Models (LLMs). For example, generating a 256 $\times$ 256 resolution image using the DiT XL/2 model can take over 17 seconds and require $10^5$ Gflops on an NVIDIA A100 GPU. Moreover, the video generation model SoRA \cite{sora}, designed concerning DiTs, contains approximately 3 billion parameters. Due to this significant parameter count, deploying them on edge devices with limited computational resources is impractical.

\begin{figure*}[!tbp]
  \centering
  \includegraphics[width=0.95\linewidth]{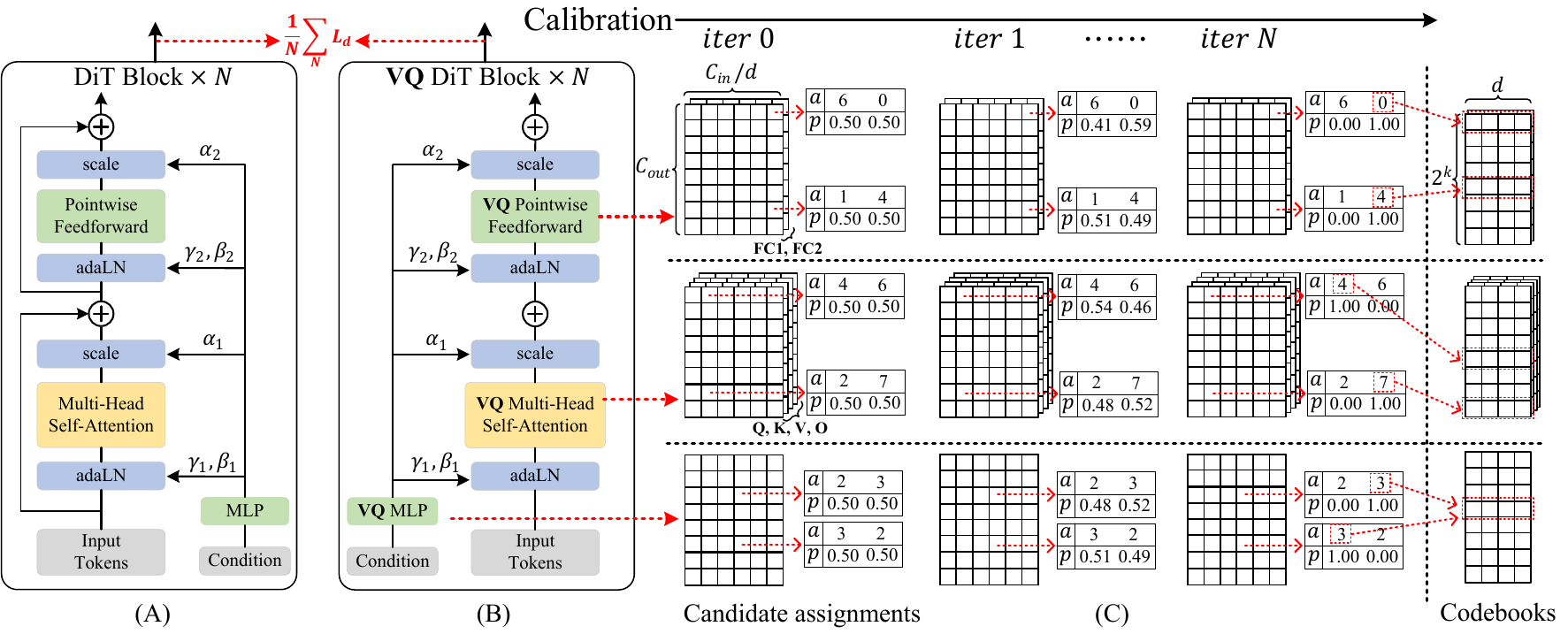}
  \caption{The pipeline of VQ4DiT. (A) DiT blocks. (B) DiT blocks are quantized by vector quantization (VQ). (C) Candidate assignments and codebooks are calibrated by zero-data and block-wise calibration to ultimately obtain the optimal assignments with the highest ratios.
  }
  \label{pipeline}
\end{figure*}

To overcome the deployment challenges, recent research has focused on the efficient deployment of diffusion models, particularly through model quantization \cite{li2023q,li2024snapfusion,he2024ptqd,wang2024quest}. Post-training quantization (PTQ) is the most widely used technique because it rapidly quantizes the original model using a small calibration set without requiring multiple iterations of fine-tuning \cite{yuan2022ptq4vit, li2023q}.  Meanwhile, vector quantization (VQ) has been shown to compress CNN models to extremely low bit-width \cite{gersho2012vector,stock2019and},  which could also be advantageous for DiTs. The classic VQ approach maps the weight sub-vectors of each layer to a codebook and assignments using clustering techniques such as the K-Means algorithm \cite{han2015deep}, and the codebook is continuously updated.

However, existing quantization methods have several limitations. First, they cannot be directly applied to DiTs, which have different network structures and algorithmic concepts compared to UNet-based diffusion models. Second, PTQ methods significantly reduce model accuracy when quantizing weights to extremely low bit-width (e.g., 2-bit). Third, traditional VQ methods only calibrate the codebook without adjusting the assignments, leading to incorrect assignment of weight sub-vectors, which provides inconsistent gradients to the codebook and ultimately results in suboptimal outcomes.

To overcome these limitations, we introduce a novel post-training vector quantization technique for the extremely low bit-width quantization of DiTs, named VQ4DiT. VQ4DiT firstly maps the weight sub-vectors of each layer to a codebook using the K-Means algorithm. It then determines a candidate assignment set for each weight sub-vector based on Euclidean distance and reconstructs the sub-vector based on the weighted average. Finally, leveraging the zero-data and block-wise calibration method, the ratio of each candidate assignment is calibrated and the optimal assignment from the set is efficiently selected while simultaneously calibrating the codebook. VQ4DiT ensures that the quantized model achieves results comparable to those of the floating-point model.
The contributions are summarized as follows:
\begin{itemize}
    \item We explore the VQ methods for extremely low bit-width DiTs and introduce DiT-specific improvements for better quantization, which have not been explored in DiT literature.
    \item We calibrate both the codebook and the assignments of each layer simultaneously, unlike traditional methods that focus solely on codebook calibration.
    \item Our method achieves competitive evaluation results compared to full-precision models on the ImageNet \cite{russakovsky2015imagenet} benchmark.
\end{itemize}

\section{Backgrounds and Related Works}
\subsection{Diffusion Transformer Models}
UNet-based diffusion models have garnered significant attention, and research has begun to explore the adoption of transformer architectures \cite{rombach2022high, croitoru2023diffusion, yang2023diffusion} within diffusion models. Recently, Diffusion Transformer Models (DiTs) \cite{peebles2023scalable} have achieved state-of-the-art performance in image generation. Notably, DiTs demonstrate scalability in terms of model size and data representation similar to large language models, making them widely applicable to image and video generation tasks \cite{sora, liu2024sora, zhu2024sora}.

DiTs consist of $N$ blocks, each containing a Multi-Head Self-Attention (MHSA) and a Pointwise Feedforward (PF) module \cite{vaswani2017attention, dosovitskiy2021image, peebles2023scalable}, both preceded by their respective adaptive Layer Norm (adaLN) \cite{perez2018film}. The structure of the DiT block is illustrated in Figure \ref{pipeline} (A). These blocks sequentially process the noised latent and conditional information, encoded as tokens in a lower-dimensional latent space \cite{rombach2022high}.

In each block, the conditional embedded information $\mathbf{c} \in \mathbb{R}^{d_{in}}$ is converted into scale and shift parameters ($\bm{\gamma}, \bm{\beta} \in \mathbb{R}^{d_{in}}$), which are regressed through MLPs and then injected into the noisy latent $\mathbf{z} \in \mathbb{R}^{n \times d_{in}}$ via adaLN:
\begin{equation}
\left\{
\begin{aligned}
(\bm{\gamma}, \bm{\beta}) &= \text{MLP}(\mathbf{c}) \\
\text{adaLN}(\mathbf{z}) &= \text{LN}(\mathbf{z}) \odot (\bm{1} + \bm{\gamma}) + \bm{\beta}
\end{aligned},
\right.
\end{equation}
where LN$(\cdot)$ denotes the Layer Norm \cite{ba2016layer}. These adaLN modules dynamically adjust the layer normalization before each MHSA and PF module, enhancing DiTs' adaptability to varying conditions and improving the generation quality.

Despite their effectiveness, DiTs demand substantial computational resources to generate high-quality images, which poses challenges to their deployment on edge devices.
In this paper, we propose an extremely low bit-width quantization method for DiTs that significantly reduces both time and memory consumption, without the need for a calibration dataset.

\begin{table}[!t]
\centering
\begin{tabular}{ccccc}
    \toprule
    Method  & $k \times d$  & $C$ (MB) & $A$ (MB) & MSE $\downarrow$ \\
    \midrule
    FP  & n/a  & n/a  & n/a & 0.00 \\
    \midrule
    3-bit UQ  & n/a  & n/a  & n/a & 7.36e-3 \\
    \midrule
    2-bit UQ  & n/a  & n/a  & n/a & 3.09e-2 \\
    \midrule
    \multirow{2}{*}{\shortstack{
    3-bit VQ}}  & $64 \times 2$ & 0.10 & 239.20 & 1.39e-3 \\
      & $4096 \times 4$ & 12.25 & 239.20 & 1.07e-3 \\
    \midrule
    \multirow{2}{*}{\shortstack{
    2-bit VQ}} & $256\times4$ & 0.77 & 159.47 & 3.98e-3 \\
      & $4096 \times 6$ & 18.38 & 159.47 & 3.35e-3 \\
    \bottomrule
\end{tabular}
\caption{Metrics of classic uniform quantization (UQ) and vector quantization (VQ) in the DiT XL/2 256$\times$256 Model. The dimensions of the codebook for VQ are represented as $k \times d$. $C$(MB) and $A$(MB) denote the memory usage of all codebooks and all assignments, respectively. 'MSE' denotes the mean square error between floating-point weights and quantized weights.}
\label{qerror}
\end{table}

\subsection{Model Quantization}
Let $ W \in \mathbb{R}^{o \times i} $ denote the weight, where $o$ represents the output channel and $i$ denotes the input channel. A standard symmetric uniform quantizer approximates the original floating-point weight $W$ as $ \widehat{W} \approx s W_{int}$, where each element in $W_{int}$ is a $b$-bit integer value and $s$ is a high-precision quantization scale shared across all elements of $W$. 

Uniform quantization and its variants of transformer blocks have been extensively studied, with most of the research focusing on the efficient quantization of model weights to reduce memory overhead. RepQ-ViT~\citep{li2023repq} adopts scale reparameterization to minimize the quantization error. GPTQ \cite{frantar2022gptq} compensates for unquantized weights based on Hessian information, achieving a good 4-bit quantization performance. Meanwhile, AWQ \cite{lin2023awq} introduces activation-aware weight quantization, specifically designed to minimize the quantization error in salient weights. Q-DiT \cite{chen2024q} employs group-wise quantization and utilizes an evolutionary search algorithm to optimize the grouping strategy.
However, uniform quantization incurs a larger error at extremely low bit-width quantization due to its limitation of reconstructing weights in equidistant distributions.

A more flexible quantization approach is vector quantization (VQ) \cite{gersho2012vector,stock2019and}, which expresses $W$ in terms of assignments $A$ and a codebook $C$.
First, VQ divides $W$ into row sub-vectors $w_{i,j} \in \mathbb{R}^{1 \times d}$:
\begin{equation}
  W =
  \begin{bmatrix}
      w_{1,1} & w_{1,2} & \cdots & w_{1,i/d} \\
      w_{2,1} & w_{2,2} & \cdots & w_{2,i/d} \\
      \vdots & \vdots & \ddots & \vdots \\
      w_{o,1} & w_{o,2} & \cdots & w_{o,i/d}
  \end{bmatrix},
\end{equation}
where $o \cdot i/d$ is the total number of sub-vectors. 
These sub-vectors are quantized to a codebook
$C = \{ c(1), \ldots, c(k) \} \subseteq \mathbb{R}^{d \times 1}$, where $c(k)$ is referred to as the $k$-th codeword.
The assignments $A = \{a_{i,j}\in \{1, \ldots, k\}\}  $ are the indices of each codeword in the codebook that best reconstruct every sub-vectors $\{ w_{i,j} \}$. 
The quantized weight $ \widehat{W} $ is reconstructed by replacing each $ w_{i,j} $ with $ c(a_{i,j}) $:
\begin{equation}
  \widehat{W} = C[A] =
  \begin{bmatrix}
      c(a_{1,1}) & c(a_{1,2}) & \cdots & c(a_{1,i/d}) \\
      c(a_{2,1}) & c(a_{2,2}) & \cdots & c(a_{2,i/d}) \\
      \vdots & \vdots & \ddots & \vdots \\
      c(a_{o,1}) & c(a_{o,2}) & \cdots & c(a_{o,i/d})
  \end{bmatrix}.
\label{cfunction}
\end{equation}
All assignments can be stored using $ \frac{o \times i}{d} \times \log_2 k $ bits and the codebook can be stored using $k \times d \times 32$ bits. To the best of our knowledge, no existing research has applied VQ to DiTs.

\begin{table}[!t]
\centering
\begin{tabular}{ccccc}
    \toprule
    Method & Fine-tune & FID $\downarrow$ & IS $\uparrow$ & Precision $\uparrow$ \\
    \midrule
    FP & n/a & 6.72  & 243.90  & 0.7848 \\
    \midrule
    3-bit UQ & Yes & 1.3e2 & 9.92 & 0.1704 \\
    \midrule
    2-bit UQ & Yes & 2.5e2 & 2.14 & 0.1081 \\
    \midrule
    \multirow{2}{*}{\shortstack{
    3-bit VQ}} & No & 46.40 & 51.86 & 0.4756 \\
     & Yes & 35.14 & 60.02 & 0.5979 \\
    \midrule
    \multirow{2}{*}{\shortstack{
    2-bit VQ}} & No & 86.82 & 18.12 & 0.3252 \\
     & Yes & 66.01 & 29.48 & 0.4533 \\
    \bottomrule
\end{tabular}
\caption{Results of classic uniform quantization (UQ) and vector quantization (VQ) in the DiT XL/2 256$\times$256 Model. 'Fine-tune' denotes whether the quantization parameters (e.g., scales or codebooks) are fine-tuned while updating the biases and normalization layers. The timesteps are set to 50 and the classifier-free guidance (CFG) is set to 1.5. The number of generated images is 10000.}
\label{fresult}
\end{table}

\section{Challenges of Vector Quantization for DiTs}
\subsection{Trade-off of codebook size} 
As illustrated in Table \ref{qerror}, we apply the classic uniform quantization (UQ) and vector quantization (VQ) to the DiT XL/2 model. At the same bit-width, VQ results in a much smaller quantization error compared to UQ. The number of codewords $k$ and their dimension $d$ significantly impact both the memory usage of VQ and the quantization error of the weights. Increasing $k$ and $d$, while keeping the memory usage of assignments constant, reduces the quantization error. However, this also increases the memory usage of the codebook, which is particularly problematic in per-layer VQ. Additionally, increasing $k$ and $d$ prolongs the runtime of the clustering algorithm and increases the subsequent calibration times. These factors necessitate a careful trade-off between quantization error and codebook size.

We utilize $k=256$ and $d=4$ for 2-bit quantization, $k=64$ and $d=2$ for 3-bit quantization. The memory usage of the codebooks is negligible when compared to the memory requirements for assignments.

\subsection{Setups of codebooks and assignments}
There are various methods to achieve VQ, one popular method being the K-Means algorithm \cite{han2015deep}. However, the quantization error of the weights can significantly degrade model performance. To mitigate the negative impact, some studies \cite{martinez2021permute,stock2019and} assume that assignments are sufficiently accurate and use the training set to fine-tune the codebook of each layer. These approaches have yielded good results on smaller CNN networks, such as ResNet18 \cite{he2016deep} and VGG16 \cite{simonyan2014very}, with performance close to that of the original models. 
Unfortunately, fine-tuning quantized DiTs on the ImageNet dataset is time-consuming and computationally intensive, while the accumulation of quantization errors is more pronounced in these large-scale models.

As shown in Table \ref{fresult}, we apply the classic UQ and VQ to the DiT XL/2 model and fine-tune the quantization parameters, ensuring identical quantization settings and updating iterations. Although VQ outperforms UQ, it still falls short of being acceptable at extremely low-width. Moreover, fine-tuning the codebook of each layer only slightly improved the results. The primary reason is that sub-vectors with the same assignment may have gradients pointing in different directions, and the accumulation of these gradients hinders the correct updating of the codeword, which results in suboptimal codewords.

Our approach differs from previous VQ methods in that we efficiently calibrate both the codebooks and the assignments simultaneously. This strategy allows us to avoid the accumulation of errors in the gradients of the codewords and to achieve better performance in DiTs.

\begin{figure*}[!t]
  \centering
  \includegraphics[width=0.9\linewidth]{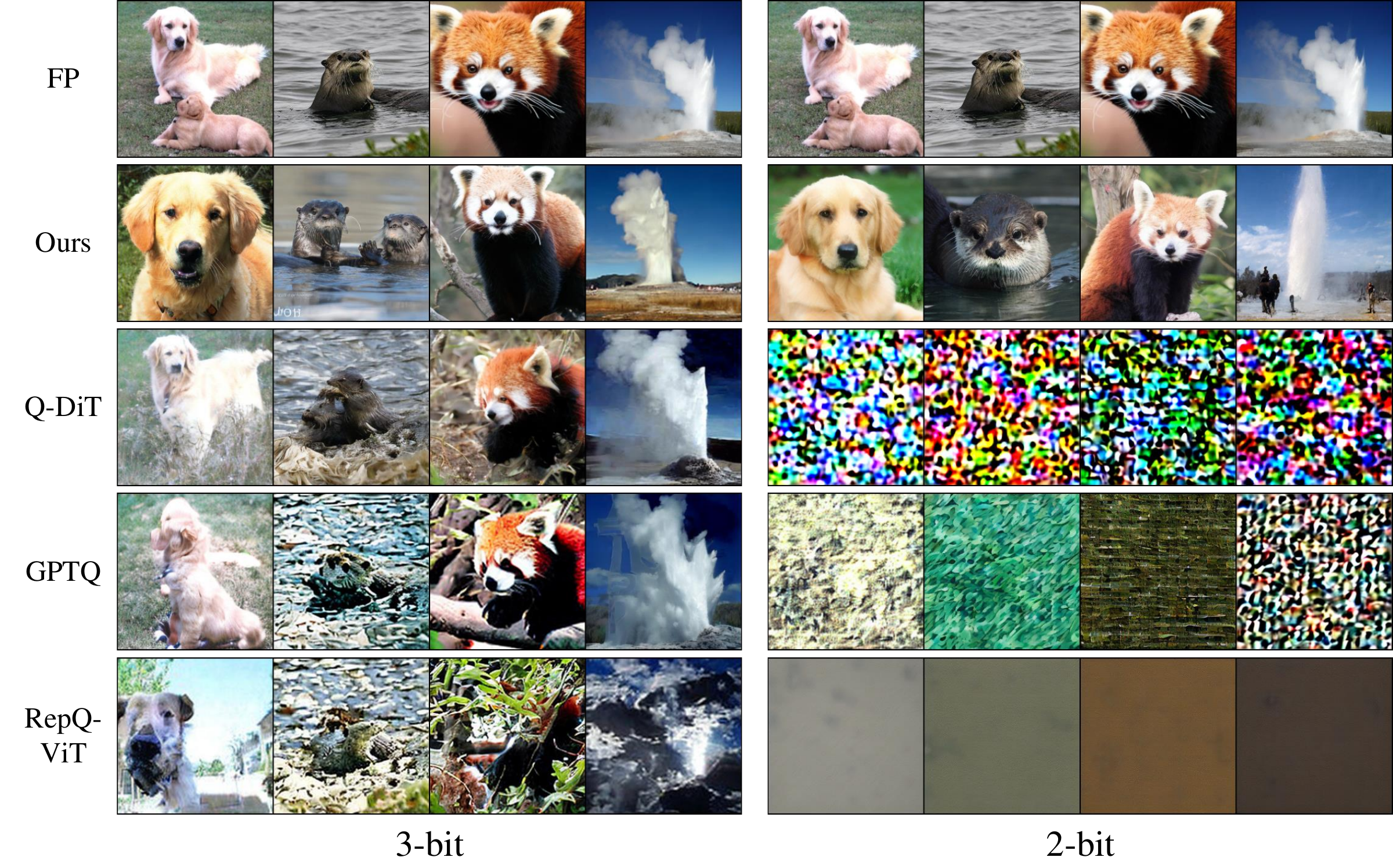}
  \caption{Images generated by VQ4DiT and three strong baselines: RepQ-ViT \cite{li2023repq}, Q-DiT \cite{chen2024q}, and GPTQ \cite{frantar2022gptq}, with 3-bit and 2-bit quantization on ImageNet 256$\times$256. Our VQ4DiT model is capable of generating high-quality images even at extremely low bit-width.
  }
  \label{image}
\end{figure*}

\section{VQ4DiT}
To address the identified challenges, we propose a novel method for efficiently and accurately vector quantizing DiTs, named Efficient Post-Training Vector Quantization for Diffusion Transformers (VQ4DiT). The description of VQ4DiT is visualized in Figure \ref{pipeline} (B) and (C). In Section 4.1, we decompose the weights of each layer of the model into a codebook and candidate assignment sets, initializing each candidate assignment with an equal ratio. In Section 4.2, we introduce a zero-data and block-wise calibration strategy to calibrate codebooks and candidate assignment sets, ultimately selecting the optimal assignments with the highest ratios.

\subsection{Initialization of Codebooks and Candidate Assignment Sets}
As shown in Equation \ref{cfunction}, the codebook $C$ and assignments $A$ of each layer can be optimized by minimizing the following objective function:
\begin{equation}
    \left\|W - C[A]\right\|_2^2 = \sum_{o,i/d} \left\|w_{o,i/d} - c(a_{o,i/d})\right\|_2^2,
\end{equation}
which can be efficiently minimized by the K-Means algorithm. However, Table \ref{fresult} demonstrates that the strategy of fine-tuning only the codebook is not effective for DiTs. Our approach considers how to calibrate both the codebook and the assignments simultaneously.

For each weight sub-vector, we calculate its Euclidean distance to all codewords, obtaining the indices of the top $n$ closest codewords:
\begin{equation}
  A_c = \{a_{o,i/d}\}_n = \arg\min_{k}^n \| w_{o,i/d} - c(k) \|_2^2,
\end{equation}
where $A_c$ is the candidate assignment sets of the sub-vectors and $n$ is the length of each $A_c$. We assume that each set contains the optimal assignment for each sub-vector, which needs to be determined. To achieve this, we assign softmax ratios $R$ to all members of the set:
\begin{equation}
  R = \{r_{o,i/d}\}_n = \{\frac{e^{z_n}}{\sum_{j=1}^n e^{z_j}}\}_n, \sum_{o,i/d} \{r_{o,i/d}\}_n = 1,
\end{equation}
where $z_n$ is the actual value of each ratio. Each ratio of $\{r_{o,i/d}\}_n$ is initialized to $\frac{1}{n}$ and calibrated in the next process. Therefore, $\widehat{W}$ can be reconstructed based on the weighted average, expressed as the formula:

\begin{equation}
    RC[A_c] = \\
    \begin{bmatrix}
        r_{1,1}c(\{a_{1,1}\}_n)  & \cdots & r_{1,i/d}c(\{a_{1,i/d}\}_n) \\
        r_{2,1}c(\{a_{2,1}\}_n)  & \cdots & r_{2,i/d}c(\{a_{2,i/d}\}_n) \\
        \vdots  & \ddots & \vdots \\
        r_{o,1}c(\{a_{o,1}\}_n) &  \cdots & r_{o,i/d}c(\{a_{o,i/d}\}_n)
    \end{bmatrix}
\end{equation}
% where $\widehat{W} = RC[A_c]$.

\subsection{Zero-data and block-wise Calibration}
Training DiTs typically relies on the ImageNet dataset \cite{russakovsky2015imagenet}. Due to its large number of images and substantial memory usage, calibrating quantized models using this dataset poses significant challenges. To more efficiently quantize DiTs, we propose a zero-data and block-wise calibration strategy, which aligns the performance of quantized models with that of floating-point models without requiring a calibration set. 

Specifically, given the same input to both the floating-point model and the quantized model, the mean square error between the outputs of each DiT block at each timestep is computed to calibrate the codebook and the ratios of the candidate assignments for each layer. It is important to note that the input for the initial timestep is Gaussian noise $\mathbf{\epsilon} \sim \mathcal{N}(0, \mathbf{I})$, and the inputs for subsequent timesteps are the outputs of the floating-point model from the previous timestep. This ensures that the quantized model does not suffer from calibration collapse due to cumulative quantization errors and that the output styles of the quantized model remain similar. Given the latent code $\mathbf{z}$ of an image and its paired conditional information $\mathbf{y} \in \{1, \ldots, 1000\}$, the block-wise calibration function is computed as:

\begin{equation}
    \mathcal{L}_d = \mathbb{E}_{\mathbf{z},\mathbf{y},d,t} \left[ \sum_{l} \left\| d_f^l(\mathbf{z}_t, \mathbf{y}, t, W) - d_q^l(\mathbf{z}_t, \mathbf{y}, t, \widehat{W}) \right\|_2^2 \right]
\end{equation}
where $\mathbf{z}_t$ represents a noisy latent at timestep $t \sim \text{Uniform}(1, T)$, and $d_fp^l(\circ)$ and $d_q^l(\circ)$ represent the $l$-th DiT block from the floating-point model and the quantized model, respectively. 

To accelerate the search for optimal assignments, we augment the mean objective function concerning $R$ as follows:
\begin{equation}
    \mathcal{L}_r =  \sum_{o,i/d,n} ( 1 - \left| 2 \times \{r_{o,i/d}\}_n - 1 \right| ) / ( \frac{o \times i}{d} ).
\end{equation}
Thus, the final objective function $\mathcal{L}$ is represented as: 
\begin{equation}
    \mathcal{L} = \lambda_d\mathcal{L}_d + \lambda_r\mathcal{L}_r,
\end{equation}
where $\lambda_d$ and $\lambda_r$ are hyperparameters, both set to 1 for simplicity. During the calibration process, we update the codebooks and ratios through gradient:
\begin{equation}
    C \leftarrow C - u \left( \pdv{\mathcal{L}}{c}, \theta \right), R \leftarrow R - u \left( \pdv{\mathcal{L}}{r}, \theta \right),
\end{equation}
where $u(\cdot, \cdot)$ is an optimizer with hyperparameters $\theta$. When $\mathcal{L}_r$ falls below a threshold $\lambda$ (e.g., $10^{-4}$), the optimal assignment for each sub-vector is the one with the highest ratio in candidate assignments, after which $R$ is no longer updated.

\begin{table*}[!t]
\centering
\begin{tabular}{cccccccc}
\toprule
Timesteps             & \begin{tabular}[c]{@{}c@{}}bit-width \end{tabular} & Method & Size (MB) & FID $\downarrow$ & sFID $\downarrow$ & IS $\uparrow$ & Precision $\uparrow$ \\ \midrule
\multirow{11}{*}{250} & 32                                   & FP     & 2553.35    & 5.33 & 17.85 & 275.13 & 0.8216 \\ \cmidrule{2-8} 
                      & \multirow{4}{*}{3}                                      & RepQ-ViT   & 239.43  & 1.5e2 & 1.3e2 & 6.73 & 0.0481 \\
                      &                                                           & GPTQ & 252.81 & 50.94 & 38.60 & 46.37 & 0.3932 \\
                      &                                                           & Q-DiT    & 249.67  & 1.3e2 & 95.86 & 12.75 & 0.1451 \\
                      &                                                           & \textbf{Ours}   & 241.14    & \textbf{10.59} & \textbf{23.03} & \textbf{267.46} &  \textbf{0.9094} \\ \cmidrule{2-8} 
                      & \multirow{4}{*}{2}                                      & RepQ-ViT   & 159.64  & 3.1e2 & 2.1e2 & 1.26 & 0.0002 \\
                      &                                                           & GPTQ & 172.86 & 2.8e2 & 1.4e2 & 3.97 & 0.0326 \\
                      &                                                           & Q-DiT    & 168.36   & 3.1e2 & 2.1e2 & 1.28 & 0.0001 \\
                      &                                                           & \textbf{Ours}   & 162.08   & \textbf{11.87} & \textbf{23.27}  & \textbf{219.33} & \textbf{0.8882} \\ \midrule
\multirow{11}{*}{100} & 32                                   & FP       & 2553.35   & 5.59 & 18.63 & 269.67 & 0.8156 \\ \cmidrule{2-8}
                      & \multirow{4}{*}{3}                                      & RepQ-ViT   & 239.43   & 1.6e2 & 1.5e2 & 4.12 & 0.0429 \\
                      &                                                           & GPTQ & 252.81 & 58.94 & 42.80 & 40.11 & 0.3529 \\
                      &                                                           & Q-DiT      & 249.67 & 1.3e2 & 1.0e2 & 11.52 & 0.1444 \\
                      &                                                           & \textbf{Ours}  & 241.14      & \textbf{10.74} & \textbf{23.90} & \textbf{265.63} & \textbf{0.9018} \\ \cmidrule{2-8} 
                      & \multirow{4}{*}{2}                                      & RepQ-ViT    & 159.64 & 3.1e2 & 2.1e2 & 1.26 & 0.0003 \\
                      &                                                           & GPTQ & 172.86 & 2.7e2 & 1.3e2 & 4.17 & 0.0001 \\
                      &                                                           & Q-DiT      & 168.36 & 3.1e2 & 2.1e2 & 1.28 & 0.0003 \\
                      &                                                           & \textbf{Ours}   & 162.08    & \textbf{11.85} & \textbf{23.64} & \textbf{213.77} & \textbf{0.8836} \\ \midrule 
\multirow{11}{*}{50}  & 32                                   & FP     & 2553.35     & 6.72 & 21.13 & 243.90 & 0.7848 \\ \cmidrule{2-8} 
                      & \multirow{4}{*}{3}                                      & RepQ-ViT & 239.43     & 1.7e2 & 1.5e2 & 3.42 & 0.0.0371 \\
                      &                                                           & GPTQ & 252.81 & 71.42 & 55.39 & 32.27 & 0.3256 \\
                      &                                                           & Q-DiT    & 249.67    & 1.5e2 & 1.2e2 & 10.13 & 0.1265 \\
                      &                                                           & \textbf{Ours}    & 241.14    & \textbf{11.91} & \textbf{24.18} & \textbf{263.93} &  \textbf{0.9096} \\ \cmidrule{2-8} 
                      & \multirow{4}{*}{2}                                      & RepQ-ViT   & 159.64  & 3.1e2 & 2.1e2 & 1.26 & 0.0003 \\
                      &                                                           & GPTQ & 172.86 & 3.1e2 & 1.5e2 & 4.33 & 0.0006 \\
                      &                                                           & Q-DiT   & 168.36    & 3.1e2 & 2.1e2 & 1.28 & 0.0005 \\
                      &                                                           & \textbf{Ours}    & 162.08   & \textbf{12.42} & \textbf{25.16} & \textbf{209.95} & \textbf{0.8725} \\ \bottomrule
\end{tabular}
\caption{Performance comparison on ImageNet 256$\times$256. 'Timesteps' denotes the sampling step of DiTs. 'bit-width' indicates the precision of quantized weights. }
\label{256_performance}
\end{table*}

\begin{table*}[!t]
\centering
\begin{tabular}{cccccccc}
\toprule
Timesteps             & \begin{tabular}[c]{@{}c@{}}bit-width \end{tabular} & Method & Size (MB) & FID $\downarrow$ & sFID $\downarrow$ & IS $\uparrow$ & Precision $\uparrow$ \\ \midrule
\multirow{8}{*}{100} & 32                                   & FP       & 2553.35   & 5.00 & 19.02 & 274.78 & 0.8149 \\ \cmidrule{2-8} 
                      & \multirow{3}{*}{3}                                      & GPTQ & 252.81 & 78.61 & 40.75 & 29.69 & 0.3604 \\
                      &                                                           & Q-DiT      & 249.67  & 2.0e2 & 1.2e3 & 5.32 & 0.0362 \\
                      &                                                           & \textbf{Ours}  & 241.14      & \textbf{33.56} & \textbf{40.95} & \textbf{67.15} & \textbf{0.7909} \\ \cmidrule{2-8} 
                      & \multirow{3}{*}{2}                                      & GPTQ & 172.86 & 3.1e2 & 1.7e2 & 2.66 & 0.0179 \\
                      &                                                           & Q-DiT      & 168.36 & 3.8e2 & 2.2e2 & 1.25 & 0.0001 \\
                      &                                                           & \textbf{Ours}   & 162.08    & \textbf{34.32} & \textbf{51.08} & \textbf{57.03} & \textbf{0.7929} \\ \midrule 
\multirow{8}{*}{50}  & 32                                   & FP     & 2553.35     & 6.02 & 21.77 & 246.24 & 0.7812 \\ \cmidrule{2-8} 
                      & \multirow{3}{*}{3}                                      & GPTQ & 252.81 & 94.35 & 52.27 & 23.96 & 0.3098 \\
                      &                                                           & Q-DiT    & 249.67    & 2.0e2 & 1.3e2 & 4.83 & 0.0317 \\
                      &                                                           & \textbf{Ours}    & 241.14    & \textbf{34.57} & \textbf{40.31} & \textbf{66.61} &  \textbf{0.7738} \\ \cmidrule{2-8} 
                      & \multirow{3}{*}{2}                                      & GPTQ & 172.86 & 3.2e2 & 1.8e2 & 2.65 & 0.0170 \\
                      &                                                           & Q-DiT   & 168.36    & 3.8e2 & 2.2e2 & 1.24 & 0.0.0001 \\
                      &                                                           & \textbf{Ours}    & 162.08   & \textbf{35.08} & \textbf{48.81} & \textbf{56.82} & \textbf{0.7744} \\ \bottomrule
\end{tabular}
\caption{Performance comparison on ImageNet 512$\times$512. 'Timesteps' denotes the sampling step of DiTs. 'bit-width' indicates the precision of quantized weights.}
\label{512_performance}
\end{table*}

\section{EXPERIMENTS}
\subsection{Experimental Settings}
\textbf{Models and quantization.} The validation setup is generally consistent with the settings used in the original DiT paper \cite{peebles2023scalable}. We select the pre-trained DiT XL/2 model as the floating-point reference model, which has two versions for generating images with resolutions of 256$\times$256 and 512$\times$512, respectively. We calibrate all quantized models using RMSprop optimizer, with a constant learning rate of $5 \times 10^{-2}$ for ratios of candidate assignments and $1 \times 10^{-4}$ for other parameters. The batch size and iteration are set to 16 and 500 respectively, allowing the experiments to be conducted on a single NVIDIA A100 GPU within 20 minutes to 5 hours. We employ a DDPM scheduler with sampling timesteps of 50, 100, and 250. The classifier-free guidance (CFG) is set to 1.5. To maintain consistency with other baselines, we only quantize the DiT blocks, which are the most computationally intensive components of the DiTs. The length of each candidate assignment set $n$ in our VQ4DiT is 2.

\textbf{Metrics.} To evaluate the quality of generated images, we follow the DiT paper and employed four metrics: Fréchet Inception Distance (FID) \cite{heusel2017gans}, spatial FID (sFID) \cite{salimans2016improved, nash2021generating}, Inception Score (IS) \cite{salimans2016improved, barratt2018note}, and Precision. All metrics were computed using ADM's TensorFlow evaluation toolkit \cite{dhariwal2021diffusion}. For both ImageNet 256$\times$256 and ImageNet 512$\times$512, we sample 10k images for evaluation.

\textbf{Baselines.} We compare VQ4DiT with three strong baselines: RepQ-ViT \cite{li2023repq}, GPTQ \cite{frantar2022gptq}, and Q-DiT \cite{chen2024q}, which are advanced post-training quantization techniques for ViTs, LLMs, and DiTs, respectively. Considering the structural similarity \cite{dosovitskiy2021image} between DiTs and the other two types of models, we re-implemented these methods and applied them to DiTs.

\subsection{Main Results} 
Tables \ref{256_performance} and \ref{512_performance} show the quantization results of the DiT XL/2 model on the ImageNet 256$\times$256 and 512$\times$512 datasets using different sample timesteps and weight bit-widths. At a resolution of 256$\times$256, our VQ4DiT achieves performance closest to that of the FP model compared to other methods. Specifically, RepQ-ViT, GPTQ, and Q-DiT undergo a significant performance drop under 3-bit quantization, which worsens as the number of timesteps decreases. In contrast, the FID increases for VQ4DiT by less than 5.3, and the IS decreases by less than 7.7. The metrics of VQ4DiT are very close to those of the FP model, indicating that our method approaches lossless 3-bit compression.

When the bit-width is reduced to 2, the other three algorithms completely collapse. VQ-DiT significantly outperforms the other three methods, with its precision decreasing by only 0.012 compared to 3-bit quantization. Figure \ref{image} shows the generated images by each algorithm, highlighting VQ4DiT’s ability to generate high-quality images even at extremely low bit-widths.

Moreover, the validation results at a resolution of 512$\times$512 mirror those at 256$\times$256, with our VQ4DiT consistently demonstrating the best performance. This indicates that VQ4DiT can generate high-quality and high-resolution images with minimal memory usage, which is crucial for deploying DiTs on edge devices.

\begin{table}[!tb]
\centering
\begin{tabular}{cccccc}
\toprule
Steps & $n$ & FID $\downarrow$  & sFID $\downarrow$     & IS $\uparrow$     & Precision $\uparrow$  \\
\midrule
\multirow{4}{*}{50} 
&  1    & 60.16             & 55.77                & 32.48             & 0.5709                \\
&  2    & 12.42            & 25.16                 & 209.95            & 0.8725                \\
&  3    & \textbf{12.22}              & \textbf{24.94}                 & \textbf{214.11}           & \textbf{0.8802}               \\
&  4    & 14.06            & 25.12                 & 176.69           & 0.8750       \\
\bottomrule
\end{tabular}
\caption{\small Ablation study on ImageNet 256$\times$256 with 2-bit quantization. $n$ denotes the length of the candidate assignment set.}
\label{ablation}
\end{table}

\subsection{Ablation Study} 
To verify the efficacy of our algorithm, we conduct an ablation study on the challenging 2-bit quantization. In Table \ref{ablation}, we evaluate the different lengths of candidate assignment sets. The detailed result indicates that as $n$ increases from 1 to 2, the performance progressively improves, validating the effectiveness of the assignment calibration. Notably, when $n = 3$, the model demonstrates the most significant performance gain, reducing the FID by 47.97 and the sFID by 30.83. However, as $n$ increases to 4, the performance worsens, suggesting that excessive candidate assignments negatively impact calibration convergence.

\begin{figure}[!t]
  \centering
  \includegraphics[width=0.9\linewidth]{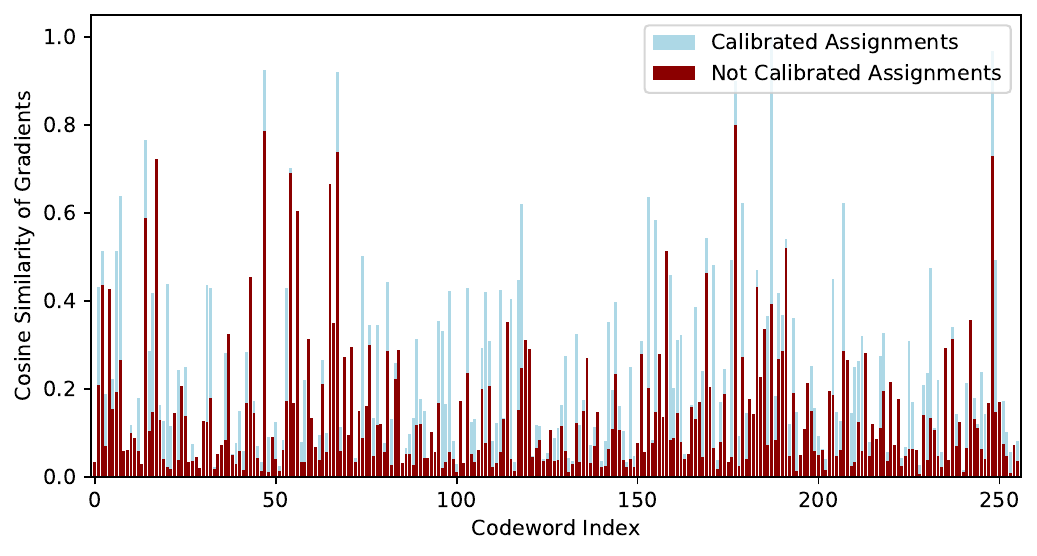}
  \caption{Cosine similarity of gradients of sub-vectors with the same assignment under the two scenarios of whether the assignments are calibrated.
  }
  \label{gradient}
\end{figure}

\begin{figure}[!t]
  \centering
  \includegraphics[width=0.9\linewidth]{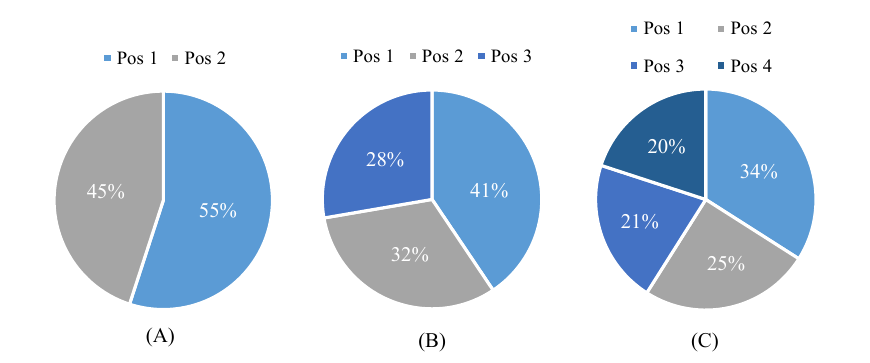}
  \caption{The proportion of position of optimal assignments in the candidate assignment sets with different lengths $n$. (A) $n=2$. (B) $n=3$. (C) $n=4$.
  }
  \label{index}
\end{figure}

To assess whether the optimal assignments yield more accurate gradients for the codebook, We calculate the gradients of the sub-vectors associated with each codeword without calibrating the codebook of each layer. As illustrated in Figure \ref{gradient}, the cosine similarity of gradients of sub-vectors with the same assignment increases significantly after the assignments are calibrated. This suggests that sub-vectors sharing the same original assignment may produce conflicting gradients for the corresponding codeword, resulting in inaccurate updates. In contrast, our assignment calibration mitigates this issue. Furthermore, as shown in Figure \ref{index}, we illustrate the distribution of the optimal assignments. Among the pool of candidate assignments, those with smaller Euclidean distances to the sub-vectors are more likely to be selected as optimal assignments.

\section{Conclusion}
In this paper, we propose a novel post-training vector quantization method, VQ4DiT, for the efficient quantization of Diffusion Transformers Models (DiTs). Our analysis identifies two main challenges when applying vector quantization (VQ) to DiTs: the need to balance the codebook size with quantization error, and the possibility that different sub-vectors with the same assignment might provide inconsistent gradient directions to the codeword. To address these challenges, we first calculate a candidate assignment set for each sub-vector. We then design a zero-data and block-wise calibration process to progressively calibrate each layer’s codebook and candidate assignment sets, ultimately leading to optimal assignments and codebooks. Experimental results demonstrate that our VQ4DiT method effectively quantizes DiT weights to 2-bit precision while maintaining high-quality image generation capabilities.

\bibliography{aaai25}

\section{Appendix}

\subsection{Additional Results}
\begin{figure*}[!t]
  \centering
  \includegraphics[width=0.735\linewidth]{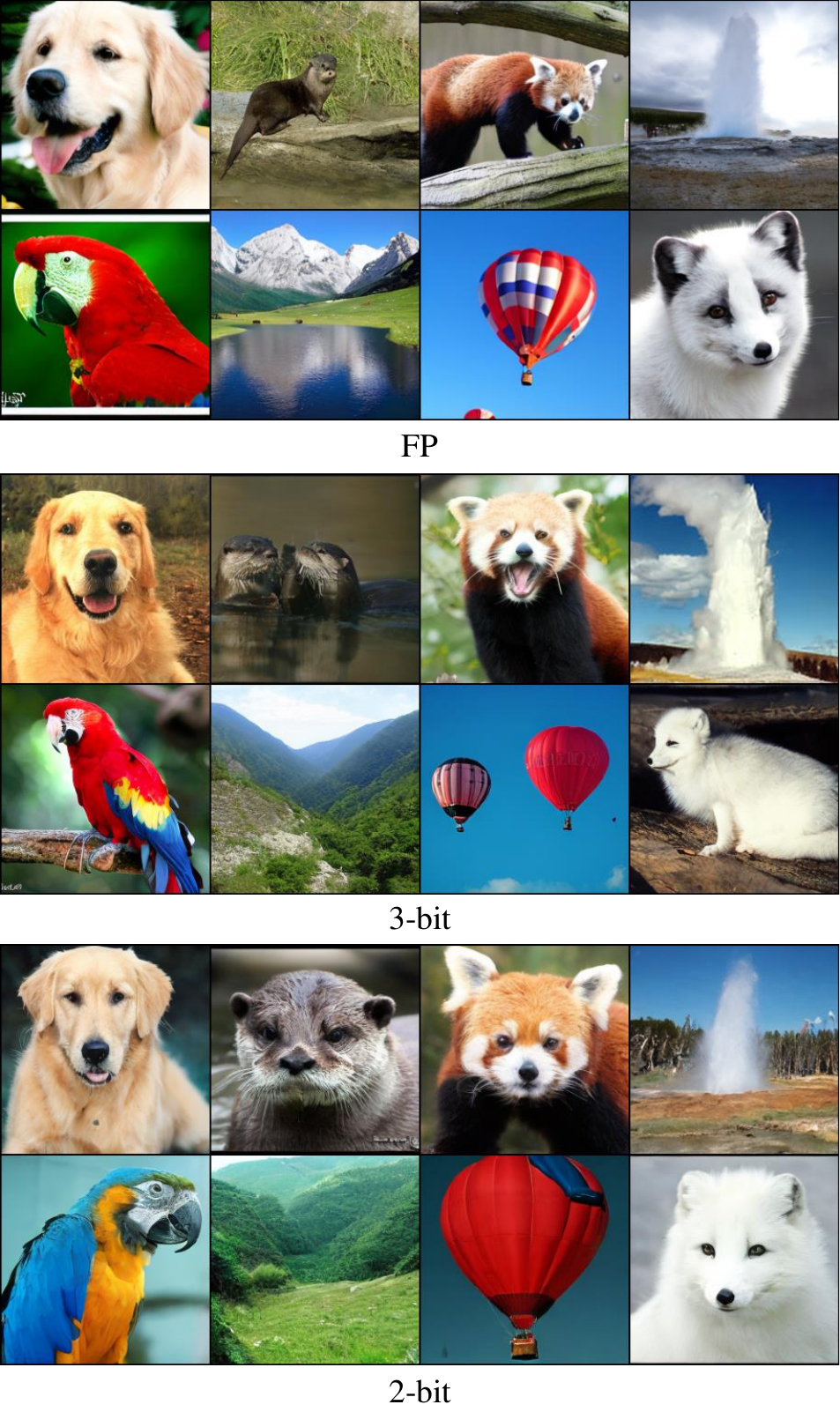}
  \caption{Images generated by VQ4DiT with 3-bit and 2-bit quantization on ImageNet 256$\times$256.
  }
  \label{256x256}
\end{figure*}

\begin{figure*}[!t]
  \centering
  \includegraphics[width=0.735\linewidth]{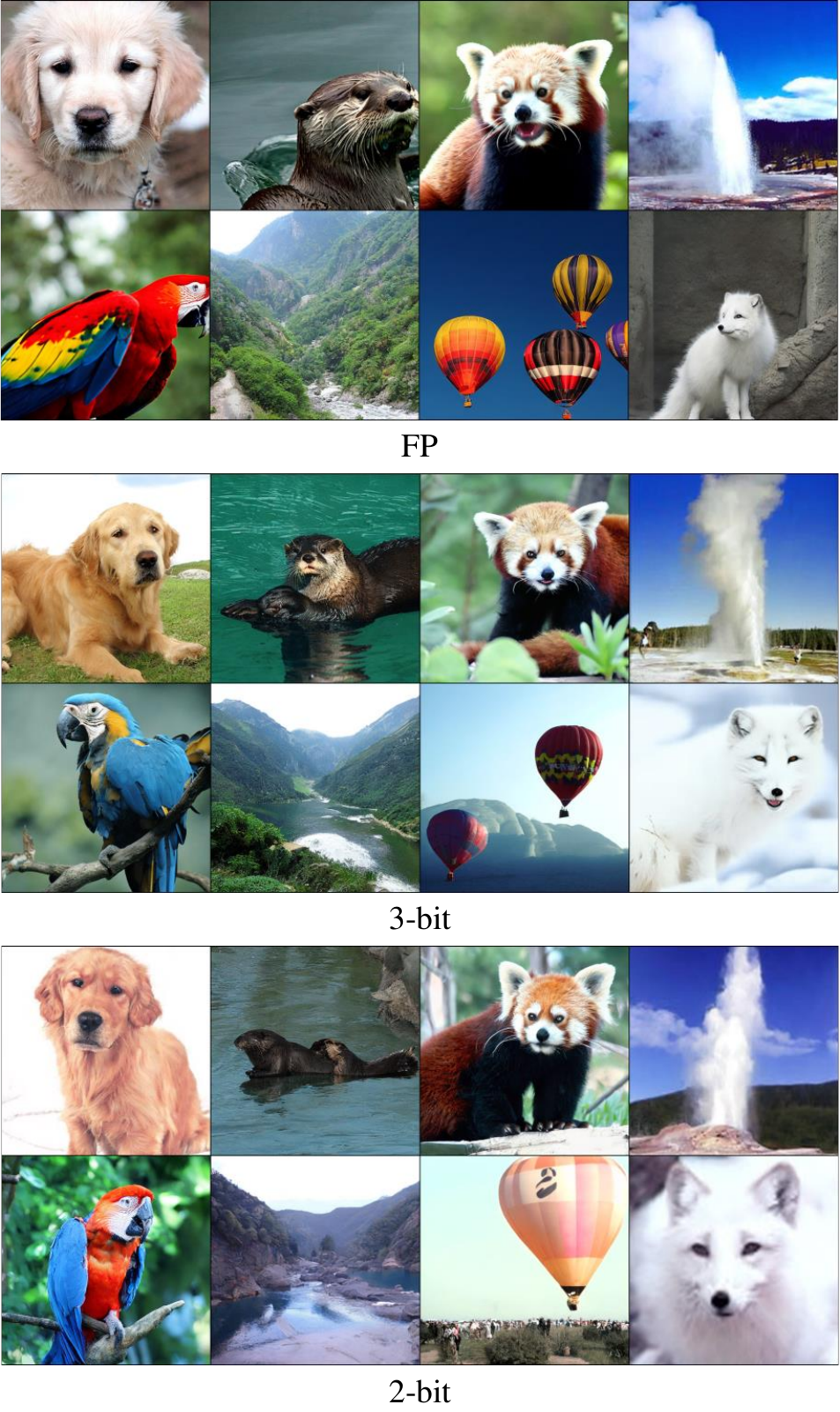}
  \caption{Images generated by VQ4DiT with 3-bit and 2-bit quantization on ImageNet 512$\times$512. 
  }
  \label{512x512}
\end{figure*}

As illustrated in Figures \ref{256x256} and \ref{512x512}, we present additional results at 256$\times$256 and 512$\times$512 resolutions. Our VQ4DiT is capable of generating high-quality images even under extremely low-bit conditions. We also present the pseudo-code of our proposed VQ4DiT in Algorithm \ref{algorithm: vqdit}.

\renewcommand{\algorithmicrequire}{\textbf{Input:}}
\renewcommand{\algorithmicensure}{\textbf{Output:}}
\begin{algorithm}
\caption{Our proposed VQ4DiT algorithm}
\label{algorithm: vqdit}
\begin{algorithmic}[1]
\REQUIRE Full-precision weight $W$ of each layer
\REQUIRE Sampling timesteps $T$, CFG scale
\REQUIRE Random conditional information $\mathbf{y} \in [1,1000]$
\ENSURE Codebook $C$ and assignments $A$ of each layer
\STATE \textbf{Initialization of Codebooks and Candidate Assignment Sets:}
\STATE Use K-Means algorithm to cluster $W$ into the initial $C$ and $A$ based on equation 4
\STATE Create candidate assignment sets $A_c$ and their ratios $R$ based on equation 5
\STATE \textbf{Zero-data and block-wise calibration:}
\FOR{$\mathbf{y}$}
    \FOR{$t=T$ to $1$}
        \STATE Generate calibration feature of each DiT block at $t$ with $W$
        \STATE Generate quantized feature of each DiT block at $t$ with $\widehat{W} = RC[A_c]$
        \STATE Calibrate and update $C$ and $R$
    \ENDFOR
    \IF{$\text{mean}(R) < 10^{-4}$}
        \STATE \textbf{break}
    \ENDIF
\ENDFOR
\STATE Select optimal assignments $A$ with largest $R$
\end{algorithmic}
\end{algorithm}

\subsection{Deployment Setup}

To improve inference speed, we implemented a CUDA vector quantization kernel for vector-vector multiplication between sub-vectors of quantized weights and sub-vectors of activations. Small-sized codebooks are loaded into shared memory to reduce bandwidth pressure. All computations are performed in FP32. As shown in Figure \ref{time}, when using the kernel, the inference time of the quantized model is reduced to approximately one-third of the original.

\begin{table}[!t]
\centering
\begin{tabular}{ccccc}
    \toprule
    Method  & Resolution & Size (MB)  & CUDA  & Time \\
    \midrule
    FP  & 256$\times$256  & 2553.35 & n/a & 61s \\
    \midrule
    \multirow{2}{*}{\shortstack{
    3-bit VQ}}  & 256$\times$256 & 241.14 & no  & 63s \\
                & 256$\times$256 & 241.14 & yes & \textbf{22s} \\
    \midrule
    \multirow{2}{*}{\shortstack{
    2-bit VQ}}  & 256$\times$256 & 162.08 & no  & 63s \\
                & 256$\times$256 & 162.08 & yes & \textbf{20s} \\
    \midrule
    \midrule
    FP  & 512$\times$512  & 2553.35 & n/a & 249s \\
    \midrule
    \multirow{2}{*}{\shortstack{
    3-bit VQ}}  & 512$\times$512 & 241.14 & no  & 253s \\
                & 512$\times$512 & 241.14 & yes & \textbf{90s} \\
    \midrule
    \multirow{2}{*}{\shortstack{
    2-bit VQ}}  & 512$\times$512 & 162.08 & no  & 252s \\
                & 512$\times$512 & 162.08 & yes & \textbf{82s} \\
    \bottomrule
\end{tabular}
\caption{Inference time(s) of DiT XL/2 on a NVIDIA A100 GPU. 'CUDA' denotes whether the CUDA vector quantization kernel is being used. During model inference, the sampling timesteps are set to 256, and the CFG scale is set to 1.5.}
\label{time}
\end{table}

\end{document}